%% file: root.tex
\documentclass[letterpaper, 10 pt, conference]{ieeeconf}  

\title{\LARGE \bf
DIV-Nav: Open-Vocabulary Spatial Relationships for \\Multi-Object Navigation
}

\author{Jesús Ortega-Peimbert, Finn Lukas Busch, Timon Homberger, Quantao Yang, and Olov Andersson}

\usepackage{graphicx}
\usepackage{booktabs}
\usepackage{url}
\usepackage{amsmath}
\usepackage{amsfonts}
\usepackage{graphicx}
\usepackage{subcaption}  
\usepackage{caption}
\usepackage{xcolor}
\usepackage{hyperref}
\usepackage{tikz}
\usepackage{pgfplots}

\usepackage{balance} 
\definecolor{jesus}{RGB}{255, 0, 0}
\definecolor{finn}{RGB}{0, 0, 255} 
\definecolor{timon}{RGB}{255, 0, 255}
\definecolor{quantao}{RGB}{255, 127, 0}
\definecolor{olov}{RGB}{30, 155, 155}

\graphicspath{{figures/}} 

\begin{document}

\maketitle
\thispagestyle{empty}
\pagestyle{empty}

\begin{abstract}

Advances in open-vocabulary semantic mapping and object navigation have enabled robots to perform an informed search of their environment for an arbitrary object. However, such zero-shot object navigation is typically designed for simple queries with an object label such as \textit{`television'} or \textit{`blue rug'}. While single- and multi-object search have progressed, real-time navigation with explicit spatial relationship reasoning during online map construction remains largely unexplored, as existing methods either rely on offline 3D reconstruction or handle only individual object targets without relational constraints. Here, we consider more complex free-text queries with spatial relationships, such as 'find the remote on the table'. We present DIV-Nav, a real-time semantic map-based navigation system for sequential multi-object search that addresses this problem through a series of relaxations: i) Decomposing natural language instructions with complex spatial constraints into simpler object-level queries, ii) computing the Intersection of individual semantic belief maps via continuous-valued scoring to identify regions where all objects co-exist, and iii) Validating discovered objects against the original spatial constraints via a vision-language model. We further investigate how to adapt frontier exploration for online semantic mapping to such spatial search queries to more effectively guide the search process. We validate our system through extensive experiments on the MultiON benchmark and real-world deployment on a Boston Dynamics Spot robot, achieving an 88\% success rate on multi-object spatial-relationship navigation tasks. More details and videos are available at https://anonsub42.github.io/reponame/.

\end{abstract}

\input{01_Introduction}

\input{02_RelatedWork}

\input{03_Method}
\input{04_ExperimentalSetup}
\input{05_Results}

\input{06_conclusion}

\balance
{\small
\bibliographystyle{IEEEtran}
\bibliography{bib}
}

\addtolength{\textheight}{-12cm}   








\end{document}

%% file: 01_Introduction.tex
\section{Introduction}
Effective robot navigation in human environments requires an understanding of natural language, where targets are often defined by their spatial relationships.
While a command like \textit{`find a chair'} only requires identifying a single object, real-world instructions often include spatial proximity constraints such as \textit{`the chair next to the desk'} or \textit{`the book on the nightstand'}. These queries are intuitive for humans, who naturally decompose them into constituent objects and reason about their arrangements. However, enabling robots to perform this reasoning in real-time during exploration remains a significant challenge.

This challenge is central to Object Goal Navigation (ObjectNav), the task where an agent must autonomously localize and navigate to a specific instance of an object category. Recent advances in Vision-Language Models (VLMs) have improved semantic understanding and 2D visual reasoning. Despite these gains, even large state-of-the-art models typically struggle with sufficient 3D understanding to robustly steer a robot over long-range navigation tasks by themselves \cite{ma2025spatialreasoner}. 

\begin{figure}[!t]
  \begin{center}
    \includegraphics[width=0.4\textwidth]{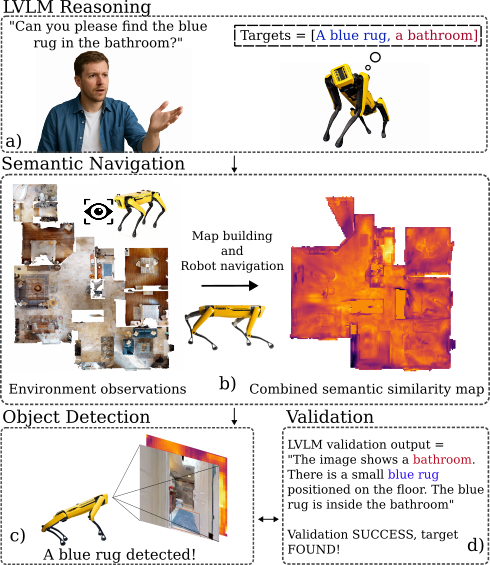}
  \end{center}
  \vspace{-0.35cm}
  \caption{Given a spatially-constrained natural language query, our system: (a) uses an LVLM to decompose the query into simpler proximity queries, (b) projects these into an online semantic belief map and intersects per-target similarity maps to localize regions where all objects likely co-exist, and (d) continuously validates candidate objects against the original constraints via an LVLM.}
  \label{fig:overview_99}
  \vspace{-0.5cm}
\end{figure}


To enable robust object navigation, current robotics research focuses mainly on hybrid approaches with conventional metric map exploration (e.g., frontier-based search) where the search objective is based on semantic similarity from a lightweight contrastive vision-language model such as CLIP \cite{radford2021learning}. Such contrastive vision-language models differ from modern large vision-language models (LVLMs) by having a simpler text encoder and only outputting a similarity score for each image and text query. Nevertheless, for simple objects these are capable of directing robotic search towards semantically promising directions (e.g., a kitchen if you are looking for a coffee machine) while retaining the robustness properties of conventional map-based navigation.

Initially, these approaches focused on single object queries \cite{yokoyama2024vlfm}. More recently, online semantic mapping has been proposed that maintains open-set semantic embedding vectors in the map, so that accumulated semantic information can be reused for multiple, consecutive search queries of single objects \cite{busch2025mapallrealtimeopenvocabulary}. However, neither of these handles queries with spatial relationship constraints, the gap DIV-Nav addresses.


We present DIV-Nav, a real-time navigation system that bridges this gap by combining LVLM-based reasoning with online semantic mapping. Our framework utilizes a three-stage process: (1) Decomposing complex instructions into simpler proximity queries via an LVLM, (2) identifying regions where objects likely co-exist through a continuous-valued Intersection operation on semantic belief maps, and (3) Validating candidate regions in image space to confirm both object presence and spatial arrangements. This loop enables real-time semantic object navigation with complex spatial relationship queries in free-text form without requiring offline 3D reconstruction while maintaining real-time performance.

The contributions of our work include:
\begin{itemize}

\item We propose a robust map-based method for online open-vocabulary object search with spatial constraints. By using an LVLM to decompose the constraints into individual object queries with relaxed proximity relationships, we show that these can be efficiently searched for online with a spatial-semantic belief map, and resulting candidates later verified against the original spatial constraint in image-space.


\item We introduce a semantic intersection operation for combining individual similarity maps from the decomposed object queries into a joint belief map, identifying regions where all queried objects are likely to co-exist. We investigate how to incorporate this combined map into the frontier exploration objective of an online semantic mapping system, guiding active exploration toward spatially relevant regions without overly constraining the search before all objects have been observed.

\item We demonstrate the system's effectiveness through comprehensive evaluation on a public benchmark and via quantitative real-world experiments with a Boston Dynamics Spot robot across four indoor environments. We make the code available to the community.\footnote{We make the code available on acceptance.}

\end{itemize}
%
%

%% file: 02_RelatedWork.tex
\section{Related Work}

\subsection{Zero-shot Object Navigation}

Visual-language navigation approaches~\cite{hong2022bridging, krantz2020beyond, wu2024vision, anderson2018vision} enable agents to follow step-by-step instructions to reach a target. Yokoyama et al.~\cite{yokoyama2024vlfm} proposed Vision-Language Frontier Maps (VLFM), a zero-shot framework that combines occupancy mapping with semantic reasoning from a pretrained VLM such as BLIP-2~\cite{li2023blip}. Their approach computes semantic similarity between real-time RGB observations and text prompts, projects the scores onto a top-down value map, and uses depth and odometry to build an occupancy map.

Huang et al.~\cite{huang2022visual, huang2025multimodal} introduced Visual Language Maps (VLMap), which fuses pretrained visual–language features with 3D reconstructions to create persistent, language-indexable spatial-semantic representations. While the resulting VLMaps allow for descriptive, spatial queries, the approach assumes the map to be pre-generated at navigation time. These works collectively illustrate the potential of combining vision-language understanding with spatial mapping and planning. More recently, Busch et al.~\cite{busch2025mapallrealtimeopenvocabulary} presented OneMap, a real-time, open-vocabulary mapping framework that constructs a belief map over CLIP-aligned features~\cite{radford2021learning}. 

Shah et al.~\cite{shah2023lm} demonstrated that composing multiple pre-trained models can enable navigation without language-annotated trajectory datasets. Similarly, Long et al.~\cite{long2024instructnav} proposed InstructNav, a unified framework capable of handling multiple instruction formats in a zero-shot setting. InstructNav’s Dynamic Chain-of-Navigation (DCoN) module translates instructions into action–landmark pairs that are continuously updated based on new observations. In contrast, DIV-Nav builds a single reusable semantic map online that can be queried consecutively for complex spatial object queries, enabling spatial relationship reasoning even on partially explored maps.


Recent efforts have addressed spatial constraints using relational graphs in MLFM \cite{raychaudhuri2025mlfmmultilayeredfeaturemaps} or multi-channel correlation maps in Finder \cite{choi2025everythinggeneralvisionlanguage}. Unlike these, DIV-Nav avoids complex graphs by employing a continuous-valued intersection of semantic belief maps to identify proximity-based co-existence. We further distinguish our work from methods in related domains: O3D-SIM \cite{Nanwani_2024} and ConceptFusion \cite{jatavallabhula2023conceptfusion} similarly resolve spatial language queries against a semantic map, but require a complete pre-built reconstruction before any queries can be answered and perform no online search planning. CORE-3D \cite{mirzaei2025core3dcontextawareopenvocabularyretrieval} follows the same offline retrieval paradigm. Additionally, LeGo-Drive \cite{paul2024legodrivelanguageenhancedgoalorientedclosedloop} addresses language-enhanced navigation for autonomous driving maneuvers, such as lane changes and parking, which differ fundamentally from the indoor object search primitives addressed in our work.

\subsection{Learned Object Navigation}
Learning-based ObjectNav has progressed via two main paths. End-to-end RL, initiated by DD-PPO~\cite{wijmansdd} and refined by modular semantic-map policies like SemExp~\cite{chaplot2020object}, has scaled to large procedural environments using powerful visual encoders~\cite{khandelwal2022simple}. Simultaneously, imitation learning reduces interaction via human demonstrations: PIRLNav~\cite{ramrakhya2023pirlnav} combines behavior cloning with RL finetuning, while newer methods utilize trajectory diffusion~\cite{yu2024trajectory} or unlabeled egocentric video~\cite{hirose24lelan}, though both necessitate substantial offline training.

A parallel effort integrates foundation models into learned pipelines. CL-CoTNav~\cite{cai2025cl} fine-tunes a VLM with chain-of-thought reasoning on expert trajectories, while Scene-LLM~\cite{fu2024scene} aligns 3D voxel representations with an LLM through multi-stage training. Others encode semantic information via implicit neural representations~\cite{marza2023multi}. These methods improve scene understanding but remain dependent on task-specific training data and fixed object vocabularies, limiting their ability to handle novel spatial relationship queries at test time.

In contrast, recent zero-shot approaches have matched or exceeded learned methods without navigation-specific training~\cite{yokoyama2024vlfm, busch2025mapallrealtimeopenvocabulary}. Our work extends this line to compositional spatial queries, a capability absent from both learned and existing zero-shot methods, while automatically benefiting from improvements in the underlying foundation models without retraining.

%% file: 03_Method.tex
\section{Method}
\label{sec:method}
\vspace{0.1cm}
\begin{figure*}[!t]
    \centering
    \vspace{6mm}
    \def\svgwidth{0.78\textwidth}
    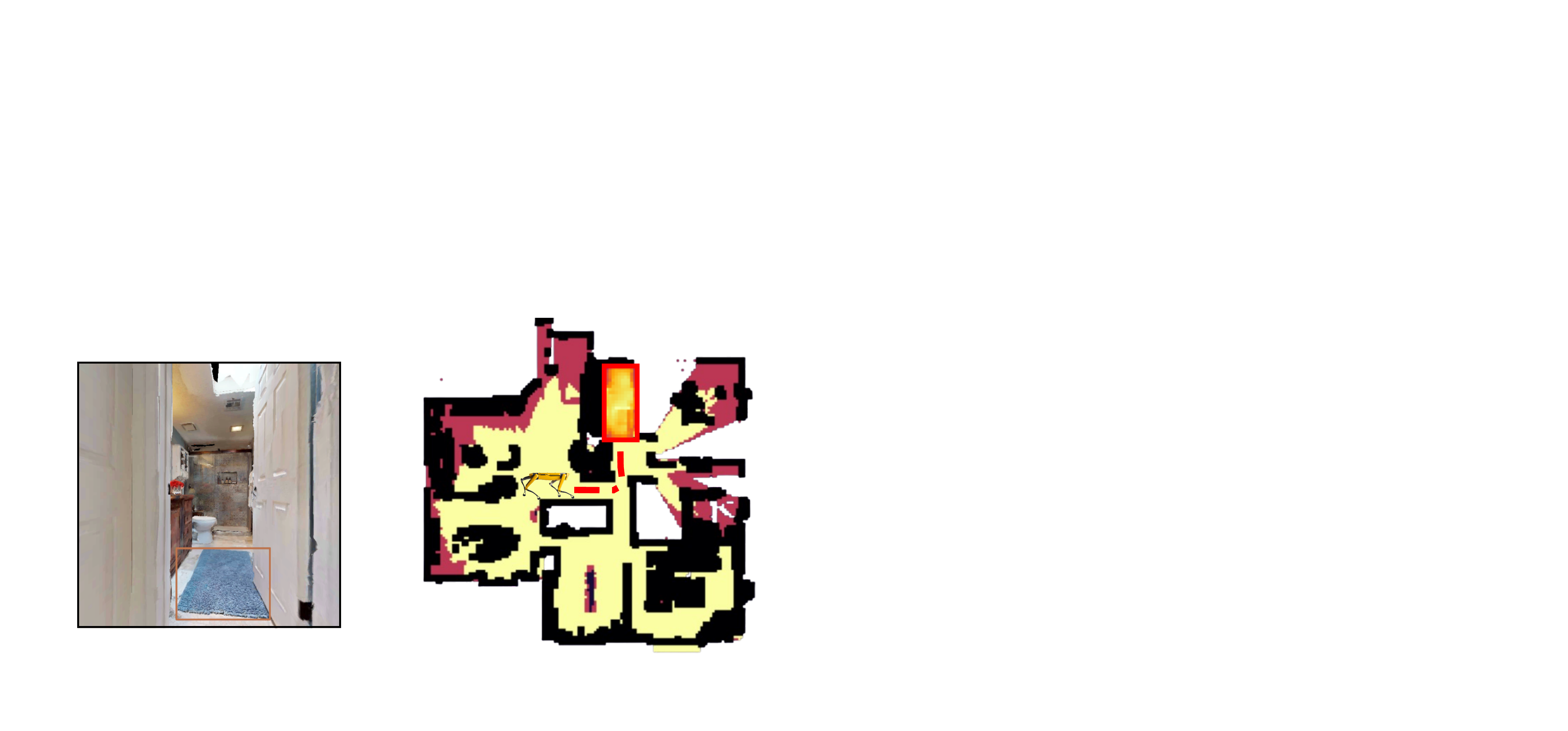
    \vspace{-0.1cm}
    \caption{Overview of DIV-Nav. (1) We \textbf{D}ecompose spatially-constrained instructions into object-level queries on a semantic map (a-b); (2) Compute Intersections to create a joint belief map that actively guides online frontier exploration toward regions where objects likely co-exist (c-d); (3) \textbf{V}alidate high-similarity regions by approaching the region, and via LVLM reasoning to confirm individual objects and their spatial relationships (e).}
    \label{fig:fig_abc}
    \vspace{-0.6cm}
\end{figure*}

Our proposed approach addresses the problem of guiding a robot towards a navigation goal that is characterized by targets with spatial relationships, inferred from a natural language input. While the traditional ObjectNav problem typically targets simple object categories, such as \textit{`a chair'}, our method enables inferring objects from abstract descriptions and searching for distinct object instances by accounting for spatial relationship constraints.

The proposed pipeline involves leveraging a compact LVLM like multimodal Phi4 5.6B \cite{microsoft2025phi4minitechnicalreportcompact} (capable of running onboard robots) to decompose the natural language inputs into multiple navigation targets that can be queried against online-constructed semantic maps, combining the resulting similarity maps to identify locations where spatial relationships are likely satisfied, and employing visual validation to confirm goal achievement.
In the following, we will refer to such models that can both process and produce natural language, as well as process visual inputs as LVLMs. Note that this differs from language-aligned vision models such as CLIP~\cite{radford2021learning} which cannot produce and process language in the same way.

\subsection{Problem Formulation}

The system is provided with a continuous stream of posed RGB-D observations $\{I_t, D_t\}$ where $I_t \in \mathbb{R}^{H \times W \times 3}$ are RGB images and $D_t \in \mathbb{R}^{H \times W}$ contains corresponding depth information. Furthermore, it is provided with an input in the form of a natural language target description $L$. $L$ either explicitly or implicitly indicates targets as well as inter-target spatial relationships. The desired output consists of navigation actions that guide the agent to the primary target that satisfies both, the target type and the spatial relationships specified in $L$.

\subsection{System Overview}

An overview of the proposed system is given in Fig.~\ref{fig:fig_abc}.
Our approach involves exploring a previously unseen environment and incrementally building a semantic representation (Fig.~\ref{fig:fig_abc} (b)), to guide a robot towards a specified navigation target. 
To this end, we adopt the uncertainty-aware semantic belief mapping and navigation framework of \cite{busch2025mapallrealtimeopenvocabulary}. This system leverages patch-level image embeddings produced by encoding the RGB frames $I_t$ using the backbone of a CLIP-aligned segmentation model ~\cite{xie2024sedsimpleencoderdecoderopenvocabulary}\cite{radford2021learning} and grounds them on a 2D grid map using depth frames $D_t$ and corresponding poses.
A Gaussian blurring kernel is applied during the integration of features to account for depth sensing noise. Querying the resulting semantic map with a text embedding yields a 2D map $S$ of similarity values.

While this similarity map provides rich spatial-semantic information, it can only be queried using relatively simple object classes that align with CLIP's feature space, limiting its applicability to complex language instructions that contain spatial relationships.
We bridge this gap by employing an LVLM as an intermediary reasoning layer that translates natural language queries into sets of basic object queries $Q$ compatible with the semantic map's CLIP-based representation (Fig.~\ref{fig:fig_abc} (a)).
The basic object queries are determined under consideration of their spatial relationships.
Hereby, we focus on \textit{proximity} relationships, which allows us to model natural language descriptions such as \textit{`inside'}, \textit{`on top of'}, \textit{`near'}, etc. Moreover, this enables us to use the LVLM to infer higher-level concepts (e.g. \textit{`towel'} $\rightarrow$ \textit{`bathroom'}), readily capturing their relationship as proximity, which can help provide guidance early on when the environment has only been sparsely explored.

Similarity maps $S_i$ that result from querying the semantic map with $q_i \in Q$ are combined to produce a joint similarity map $S_\mathrm{comb}$ (Fig.~\ref{fig:fig_abc} (c)).
High-similarity regions in $S_\mathrm{comb}$ indicate locations where the targets corresponding to $q_i$ are likely located close to each other and are used to guide the navigation module (Fig.~\ref{fig:fig_abc} (d)) inherited from \cite{busch2025mapallrealtimeopenvocabulary}. Crucially, rather than treating this as a static search, our framework incorporates $S_\mathrm{comb}$ into the online frontier exploration objective. By weighting unexplored boundaries (frontiers) with these similarity scores, the system actively biases the search process toward spatially relevant regions even before all objects have been explicitly observed in the environment.
Upon reaching such a location, we utilize the LVLM to validate discovered objects against the original spatial constraint specifications in $L$ (Fig.~\ref{fig:fig_abc} (e)).
If the validation does not confirm object presence or spatial relation, the search continues.

%
\subsection{Decomposing Natural Language Queries}
\label{sec:decomposition}
To decompose a given natural language query into object and location targets for which the semantic map can be queried, we prompt the LVLM to reason over the given command and extract the following information:
\begin{enumerate}
    \item Identify all objects and locations explicitly mentioned in the instructions, $T_\mathrm{expl}$.
    \item Infer reasonable higher-level concepts or locations if not explicitly stated, $T_\mathrm{inferred}$. For \textit{`a towel'}, the LVLM might infer \textit{`kitchen'} or \textit{`bathroom'}.
    \item Infer possible objects if the prompt is a demand rather than an explicit target $T_\mathrm{impl}$. For \textit{`the room is on fire!'}, the LVLM extracts \textit{`fire extinguisher'}.
    \item Identify the primary target $\hat{T}$, and the spatial relationship of all other extracted targets with the primary target, $R(\cdot, \hat{T})$. For \textit{`the blue rug in the bathroom'}, \textit{`rug'} becomes the primary target, and the relationship is \textit{`in'}.
\end{enumerate}
As a result, we obtain a primary target $\hat{T}$, and a list of other relevant targets $T_{\text{all}} = \hat{T} \cup T_{\text{expl}} \cup T_{\text{inferred}} \cup T_{\text{impl}}$.
We further filter this list to retain only targets $t \in T_{\text{all}}$ where $R(t, \hat{T})$ can be understood as a proximity relationship.
For instance, we retain proximity relationships such as \textit{`in'}, \textit{`on top'}, \textit{`near'},  but discard non-proximity relationships such as \textit{`not in'}, \textit{`far from'}.
As a result, for \textit{`the rug \textbf{not} in the bathroom'}, we would discard \textit{`bathroom'} from the list.
We denote the resulting filtered target set $Q$.

\subsection{Querying the Map}
\label{sec:method_query}
Given the decomposed target set $Q$, the goal of this module is to identify map regions where all constituent objects are likely to co-exist in physical space. Each query $q_i \in Q$ is encoded with the CLIP~\cite{radford2021learning} text encoder to obtain an embedding $q_{\mathrm{CLIP},i} \in \mathbb{R}^f$, and its cosine similarity against every map cell yields a per-query similarity map $S_i(x, y)$.

To aggregate these individual maps into a single \emph{spatial co-existence score}, we require a score that is high only where \emph{all} objects are simultaneously likely to be present, which is the defining condition for proximity relationships such as \textit{`on'}, \textit{`inside'}, and \textit{`next to'}. We formulate this as a continuous-valued intersection, i.e. $S_\mathrm{int} = \min_i S_i(x,y)$.
The minimum has a clear geometric interpretation: a cell scores highly only if every constituent object has high semantic similarity there, and is suppressed as soon as any one object is unlikely to be present. This is strictly tighter than alternatives such as a sum or average, which can yield high scores even when individual object matches are weak. Crucially, the operation is applied over \emph{continuous} similarity values rather than binary thresholds, which avoids per-object threshold tuning in 2D map space and preserves graded navigation guidance throughout exploration.

A useful property of this formulation is its interaction with the spatial blurring inherent to the semantic map, which applies a Gaussian kernel during feature integration to account for depth noise and FOV boundary uncertainty. This blurring causes similarity responses for spatially adjacent objects to overlap, so $S_\mathrm{int}$ correctly highlights their shared vicinity even for relationships like \textit{`next to'} that do not correspond to a strict geometric intersection.

At early stages of exploration, unobserved targets produce near-uniform low similarity across the map, which would suppress $S_\mathrm{int}$ and leave the robot without directional guidance. We address this by blending the intersection with the strongest individual match:
\begin{align}
S_{\text{comb}}(x,y) = \alpha \cdot S_{\text{int}}(x,y) + (1 - \alpha) \cdot \max_i S_i(x,y),
\label{eq:scomb}
\end{align}
where $\alpha$ = 0.9 is fixed throughout our navigation, strongly prioritizing $S_\mathrm{int}$ once sufficient coverage exists, while the $\max$ term helps the robot toward the most semantically relevant observed region during sparse early exploration. For queries where no additional context targets are extracted, $Q = \{\hat{T}\}$ and $S_\mathrm{comb}$ reduces to $S_{\hat{T}}$, recovering standard single-object behavior. An example of the decomposition and resulting similarity maps is shown in Fig.~\ref{fig:fig_abc}.

\subsection{Navigation and Validation}
\label{sec:method_nav}

The navigation strategy, including the map representations and frontier-scoring mechanism, is adopted directly from \cite{busch2025mapallrealtimeopenvocabulary} and extended here to operate on the multi-object combined map $S_\mathrm{comb}$, which directs the robot toward regions most likely to satisfy the full spatial arrangement specified in the query. The system maintains four binary maps derived from the uncertainty estimates of the semantic map: the \textbf{Observed Map} $\mathcal{O}$ (regions reached by at least one sensor update), the \textbf{Semantically Explored Map} $\mathcal{E}$ (regions where uncertainty falls below 0.2, indicating reliable semantic features), the \textbf{Searched Map} $\mathcal{C}$ (same threshold, but for the current query; reset at each new query), and the \textbf{Navigable Map} $\mathcal{N}$ (obstacle-free regions available for path planning).

Navigation candidates are drawn from two sources. \textbf{Frontier candidates} lie on the boundary between $\mathcal{E}$ and $\mathcal{O}$, representing areas observed but not yet semantically reliable; each frontier is scored by $\max(S_\mathrm{comb}(x,y))$ over reachable cells in $\mathcal{O} - \mathcal{E}$. \textbf{Cluster candidates} are high $S_\mathrm{comb}$ regions within ($\mathcal{E} - \mathcal{C}$), representing previously explored areas that warrant revisiting under the current query. Each cluster is scored by its peak $S_\mathrm{comb}$ value.

A key property of this frontier scoring is that $S_\mathrm{comb}$, through its intersection term, biases exploration toward regions where \emph{all} decomposed sub-targets are likely to co-exist. This steers the robot away from isolated instances of any single object and toward the specific location where the full spatial relationship can be verified, a behavior that single-object navigation scores cannot produce.

The robot greedily selects the highest-scoring candidate from either source, and A* on $\mathcal{N}$ computes a collision-free path. During navigation, object detections are gated by a consensus filter: a candidate is accepted only when the detector fires \emph{and} the current cell falls in the top 5-percentile of $S_\mathrm{comb}$, suppressing detections of objects that match the target type but are in the wrong spatial context.

Upon a passing detection, the robot approaches within $0.5\,\mathrm{m}$ and triggers LVLM validation, prompting it to confirm (1) that the primary target is present and (2) that the original spatial constraint is satisfied. If the LVLM confirms that the conditions are satisfied, we terminate the search. We reset the searched map, and receive a new natural language query if available.

The Decompose-Intersect-Validate framework addresses three key challenges in spatial relationship navigation. First, 
decomposition via LVLM enables translation from complex natural language (which CLIP cannot directly encode) into simpler object-level queries 
compatible with semantic maps. Second, continuous-valued intersection provides spatial reasoning without requiring explicit 3D instance 
segmentation or object pose estimation—high-scoring regions in the intersection map indicate where multiple object beliefs overlap, 
suggesting spatial proximity. Third, LVLM validation is necessary because semantic maps may have high similarity scores for objects that match the 
type but not the specific instance or spatial arrangement (e.g., multiple tables in a room). Together, these components enable efficient spatial 
search (via map-based reasoning) while maintaining accuracy (via visual validation).

%% file: figures/figure_2_iros.pdf_tex
\begingroup%
  \makeatletter%
  \providecommand\color[2][]{%
    \errmessage{(Inkscape) Color is used for the text in Inkscape, but the package 'color.sty' is not loaded}%
    \renewcommand\color[2][]{}%
  }%
  \providecommand\transparent[1]{%
    \errmessage{(Inkscape) Transparency is used (non-zero) for the text in Inkscape, but the package 'transparent.sty' is not loaded}%
    \renewcommand\transparent[1]{}%
  }%
  \providecommand\rotatebox[2]{#2}%
  \newcommand*\fsize{\dimexpr\f@size pt\relax}%
  \newcommand*\lineheight[1]{\fontsize{\fsize}{#1\fsize}\selectfont}%
  \ifx\svgwidth\undefined%
    \setlength{\unitlength}{2349.43993756bp}%
    \ifx\svgscale\undefined%
      \relax%
    \else%
      \setlength{\unitlength}{\unitlength * \real{\svgscale}}%
    \fi%
  \else%
    \setlength{\unitlength}{\svgwidth}%
  \fi%
  \global\let\svgwidth\undefined%
  \global\let\svgscale\undefined%
  \makeatother%
  \begin{picture}(1,0.46648962)%
    \lineheight{1}%
    \setlength\tabcolsep{0pt}%
    \put(0,0){\includegraphics[width=\unitlength,page=1]{figure_2_iros.pdf}}%
    \put(0.95126575,0.45522648){\color[rgb]{0,0,0}\makebox(0,0)[t]{\lineheight{1.25}\smash{\begin{tabular}[t]{c}\textbf{I}ntersect\end{tabular}}}}%
    \put(0.06287432,0.02317845){\color[rgb]{0,0,0}\makebox(0,0)[t]{\lineheight{1.25}\smash{\begin{tabular}[t]{c}\textbf{V}alidate\end{tabular}}}}%
    \put(0,0){\includegraphics[width=\unitlength,page=2]{figure_2_iros.pdf}}%
    \put(0.90096303,0.23405444){\color[rgb]{0.50196078,0.50196078,0.50196078}\makebox(0,0)[t]{\lineheight{1.25}\smash{\begin{tabular}[t]{c}\small"a  bathroom"\end{tabular}}}}%
    \put(0.90970403,0.0292934){\color[rgb]{0.50196078,0.50196078,0.50196078}\makebox(0,0)[t]{\lineheight{1.25}\smash{\begin{tabular}[t]{c}\small "a blue rug"\end{tabular}}}}%
    \put(0.68881895,0.01676833){\color[rgb]{0.50196078,0.50196078,0.50196078}\makebox(0,0)[t]{\lineheight{1.25}\smash{\begin{tabular}[t]{c}\small combined map $S_\mathrm{comb}$\end{tabular}}}}%
    \put(0,0){\includegraphics[width=\unitlength,page=3]{figure_2_iros.pdf}}%
    \put(0.29205261,0.36436679){\makebox(0,0)[t]{\lineheight{1.25}\smash{\begin{tabular}[t]{c}LVLM\end{tabular}}}}%
    \put(0.02758095,0.39725767){\color[rgb]{0,0,0}\makebox(0,0)[lt]{\lineheight{1.25}\smash{\begin{tabular}[t]{l}"Find the blue\\rug inside the\\bathroom!"\end{tabular}}}}%
    \put(0.45793092,0.44854706){\color[rgb]{0,0,0}\makebox(0,0)[t]{\lineheight{1.25}\smash{\begin{tabular}[t]{c}\textbf{D}ecompose\end{tabular}}}}%
    \put(0,0){\includegraphics[width=\unitlength,page=4]{figure_2_iros.pdf}}%
    \put(0.66105674,0.3824241){\color[rgb]{0.10196078,0.10196078,0.10196078}\makebox(0,0)[t]{\lineheight{1.25}\smash{\begin{tabular}[t]{c}Semantic\\Map\end{tabular}}}}%
    \put(0,0){\includegraphics[width=\unitlength,page=5]{figure_2_iros.pdf}}%
    \put(0.40073616,0.38803805){\color[rgb]{0,0,0}\makebox(0,0)[lt]{\lineheight{1.25}\smash{\begin{tabular}[t]{l}"a blue rug",\\"a bathroom"\end{tabular}}}}%
    \put(0,0){\includegraphics[width=\unitlength,page=6]{figure_2_iros.pdf}}%
    \put(0.39666841,0.03101987){\color[rgb]{0.50196078,0.50196078,0.50196078}\makebox(0,0)[t]{\lineheight{1.25}\smash{\begin{tabular}[t]{c}Navigate\end{tabular}}}}%
    \put(0,0){\includegraphics[width=\unitlength,page=7]{figure_2_iros.pdf}}%
    \put(0.69670724,0.34307953){\color[rgb]{0,0,0}\makebox(0,0)[lt]{\lineheight{1.25}\smash{\begin{tabular}[t]{l}\small b)\end{tabular}}}}%
    \put(0.31072409,0.32228444){\color[rgb]{0,0,0}\makebox(0,0)[lt]{\lineheight{1.25}\smash{\begin{tabular}[t]{l}\small a)\end{tabular}}}}%
    \put(0.79154607,0.41739659){\color[rgb]{0,0,0}\makebox(0,0)[lt]{\lineheight{1.25}\smash{\begin{tabular}[t]{l}\small c)\end{tabular}}}}%
    \put(0.47793474,0.25001786){\color[rgb]{0,0,0}\makebox(0,0)[lt]{\lineheight{1.25}\smash{\begin{tabular}[t]{l}\small d)\end{tabular}}}}%
    \put(0.05390997,0.25064203){\color[rgb]{0,0,0}\makebox(0,0)[lt]{\lineheight{1.25}\smash{\begin{tabular}[t]{l}\small e)\end{tabular}}}}%
  \end{picture}%
\endgroup%

%% file: 04_ExperimentalSetup.tex
\vspace{-0.2cm}
\section{Experimental Setup}

We evaluate our method for multi-object navigation tasks given in natural language in the MultiON Challenge 2024 benchmark~\cite{multion2024}, as well as through real-robot navigation experiments. We use Phi-4-multimodal-instruct \cite{microsoft2025phi4minitechnicalreportcompact} as our LVLM model. For object detection, see Sec.~\ref{sec:method_nav}, we follow \cite{busch2025mapallrealtimeopenvocabulary} and use YOLOv7~\cite{yolov7} for MS-COCO~\cite{mscoco} classes, and an open-set detector Yolo-World~\cite{yoloworld} otherwise.
%


%
\textbf{Multi-Object Navigation in HSSD:} 
The MultiON 2024 challenge extends the traditional object navigation framework presented in \cite{wani2020multion} by including navigation targets described by language instructions, such as \textit{`find the red short pillar candle on the grey nightstand'}. Each of these language instructions may contain fine-grained descriptions of the target (e.g. \textit{`find the mini spa candle'}) or may also contain spatial relations between objects, such as \textit{`the mantel clock on the chest of drawers'}.

The task requires agents to navigate to a sequence of three objects located within realistic 3D environments 
from the Habitat Synthetic Scenes Dataset (HSSD) \cite{habitat19iccv}. Each episode begins with the agent at a random starting position and orientation within an unseen environment. The agent receives a sequence of three natural language descriptions corresponding to target objects that must be found in sequence, but is only informed about the subsequent target once the previous has been found. Navigation is considered successful when the agent calls the FOUND action within $0.5\,\mathrm{m}$ of each target object. The episode terminates either when all objects are successfully found or when an incorrect FOUND action is called. We evaluate on the minival split of the MultiON challenge, which consists of 100 episodes with a sequence of three target objects in 20 scenes.
%
%
%
%
%
The agent has access to RGB-D camera observations, providing 256×256 resolution images with a $79^\circ$\ horizontal field of view, and a noiseless GPS+Compass sensor that provides location and orientation of the agent relative to the agent's initial position at episode start. The action space consists of discrete navigation primitives: move forward $0.25\,\mathrm{m}$, turn left/right $15^\circ$\, and the FOUND action to indicate object discovery. 

\textbf{Multi-Object Navigation in the Real World:}
We deploy our DIV-Nav system on a Boston Dynamics Spot quadruped robot. For observations, the system uses a single front-facing RealSense D455 stereo depth camera and a Livox Mid 360 lidar running Fast-LIO2 \cite{DBLP:journals/corr/abs-2107-06829} for odometry. While LiDAR is utilized here to ensure robust odometry via Fast-LIO2, we emphasize that the DIV-Nav framework is sensor-agnostic regarding pose estimation. The choice of odometry source affects the global consistency of the metric map but does not alter the fundamental algorithmic approach to semantic mapping or spatial relationship reasoning.
For real-world experiments, we adapt the feature localization parameters of the mapping to account for depth noise.
Except for GPT-4o-mini, we run the entire stack on-board a Jetson Orin AGX, with the robot following a path-tracking controller publishing standard velocity commands. The full stack runs at 1.6 Hz map update rate; LVLM decomposition averages 1.046 s and LVLM validation averages 1.757 s per candidate, both infrequent relative to the map update cycle.

Inspired by the Multi-On challenge episode setup, we conduct 15 Multi-Object navigation experiments across four different scenes: an office waiting room, a robotics lab, a kitchen area, and an office lounge. The selected real-world scenes encompass significant variety in layout and illumination, ranging from naturally lit office lounges to artificially lit laboratory environments. This variety allows us to evaluate the robustness of the underlying CLIP-based semantic features to lighting variations and structural differences, such as open floor plans and narrow corridors. The targets cover both simple objects and more complex target descriptions with spatial relationships. In contrast to the MultiON benchmark, we do not terminate an episode if the robot incorrectly identifies a target and allow it to always attempt all three targets per episode. Note that we report the success rate (SR) \textit{per object}.

%

\textbf{Metrics}: For multi-object navigation, we report \textit{Progress} (Pr), the average fraction of found objects from the total number of targets per episode, as well as \textit{Success} (SR), the percentage of episodes where the agent successfully finds all target objects in the sequence.  We further define \textit{Success Rate per Attempted Target} (SRAT) as the success rate only for attempted object goals. 

\textbf{Baselines}: We compare against the two MultiON challenge submissions, i.e. MOPA \cite{raychaudhuri2024mopa} and IntelliGO Labs \cite{multion2024} which were trained for the task using RL.

We include~\cite{busch2025mapallrealtimeopenvocabulary} as 
an additional baseline to isolate our spatial reasoning 
contribution; since it handles only simple object queries, 
we provide it with the primary target via our decomposition 
module (Sec.~~\ref{sec:decomposition}).

To characterize query complexity, we define three dimensions: (1) the number of target objects extracted by the decomposition step, (2) the number of spatial relationships present, and (3) the relationship type: proximity/containment (e.g., \textit{`on'}, \textit{`in'}, \textit{`near'}) versus non-proximity (e.g., \textit{`far from'}, \textit{`not in'}). Notably, fine-grained single-object descriptions without relational structure challenge CLIP's feature space rather than the spatial reasoning pipeline, and represent a distinct failure mode from relational queries.

The map covers a 60×60 m area discretized into a uniform grid of 600×600 cells (0.1 m/cell) for simulation, and 300×300 cells (0.2 m/cell) for real-world deployment.

%% file: 05_Results.tex
\section{Results}
Our experiments were designed to answer two key questions: (1) Can our semantic intersection approach handle spatially-constrained navigation queries better than traditional zero-shot object-navigation approaches? (2) Can our method be successfully deployed on a real robot for natural language command search in real environments?
\subsection{Multi-Object Navigation in HSSD}
We evaluate DIV-Nav against several baselines on the MultiON challenge minival split. 
Table~\ref{tab:cf_results_asr_progress_sr} presents our results compared to MOPA~\cite{raychaudhuri2024mopa}, IntelliGO Labs (the winning submission from the 2024 challenge)~\cite{multion2024}, and OneMap~\cite{busch2025mapallrealtimeopenvocabulary} provided with the primary target.

\begin{table}[h]
\centering
\begin{tabular}{l c c c | c c}
\toprule
Approach & SRAT $\uparrow$ & Pr $\uparrow$ & SR $\uparrow$ & FP ($\downarrow$) & TNF ($\downarrow$)\\
\midrule
MOPA~\cite{raychaudhuri2024mopa} & 0.05 & 0.02 & 0.00 & - & -\\
IntelliGO Labs~\cite{multion2024} & 0.23 & 0.10 & 0.03 & - & -\\
OneMap~\cite{busch2025mapallrealtimeopenvocabulary} & 0.24 & 0.10 & 0.03 & 0.76 & 0.0\\
DIV-Nav (Ours) & \textbf{0.30} & \textbf{0.14} & \textbf{0.04} & 0.44 & 0.25\\
\bottomrule
\end{tabular}
\caption{Comparison of navigation performance across Success Rate per Attempted Target (SRAT), Progress (Pr), and Success Rate (SR). To further understand different failure cases, we report the False Positive Rate (FP), where the agent called the FOUND action on a wrong target, and the Terminate-Not-Found Rate (TNF) where the agent terminates the search without calling the found action.}
\label{tab:cf_results_asr_progress_sr}
\end{table}

DIV-Nav achieves the best performance across all metrics, with a 30\% improvement in SRAT over the baselines and a 40\% improvement in Progress. While absolute performance remains challenging due to the complexity of the benchmark, our approach demonstrates advantages in handling spatially-constrained queries.
We further observe that our method reduces false positives compared to \cite{busch2025mapallrealtimeopenvocabulary}, which we attribute to our method being able to reason about spatial constraints, instead of just searching for the primary target.
In contrast to that, \cite{busch2025mapallrealtimeopenvocabulary} might call the FOUND action on a target that matches the primary target, but does not satisfy the full constraint.

The relatively low absolute performance across all methods can be attributed to several factors: (1) Low-resolution observations: The 256×256 RGB-D images limit detailed object recognition; (2) LVLM performance in simulation: Vision-language models show reduced effectiveness on the simple graphics used in the benchmark compared to real-world images; (3) Highly challenging queries: Using our complexity characterization from Section IV, the benchmark's hardest queries are predominantly fine-grained single-object descriptions (one target, zero spatial relationships) such as \textit{`Find the freestanding bath/shower mixer'} and \textit{`find the cream chair by Joanna Gaines'}, which challenge CLIP's fine-grained discrimination capacity rather than the spatial reasoning pipeline, and represent limitations of the underlying semantic map rather than the DIV-Nav framework itself. Lastly, our semantic map struggles with small or occluded objects in the feature space.

Despite these challenges, our semantic intersection approach shows consistent improvements over baselines, validating our core hypothesis that decomposing complex spatial queries and combining similarity maps leads to better navigation performance.
%
%

\subsection{Real World Experiments}

\begin{figure}[ht]
    \centering
    \vspace{0.2cm}
    \begin{subfigure}[t]{0.2\textwidth}
        \centering
        \includegraphics[width=\textwidth]{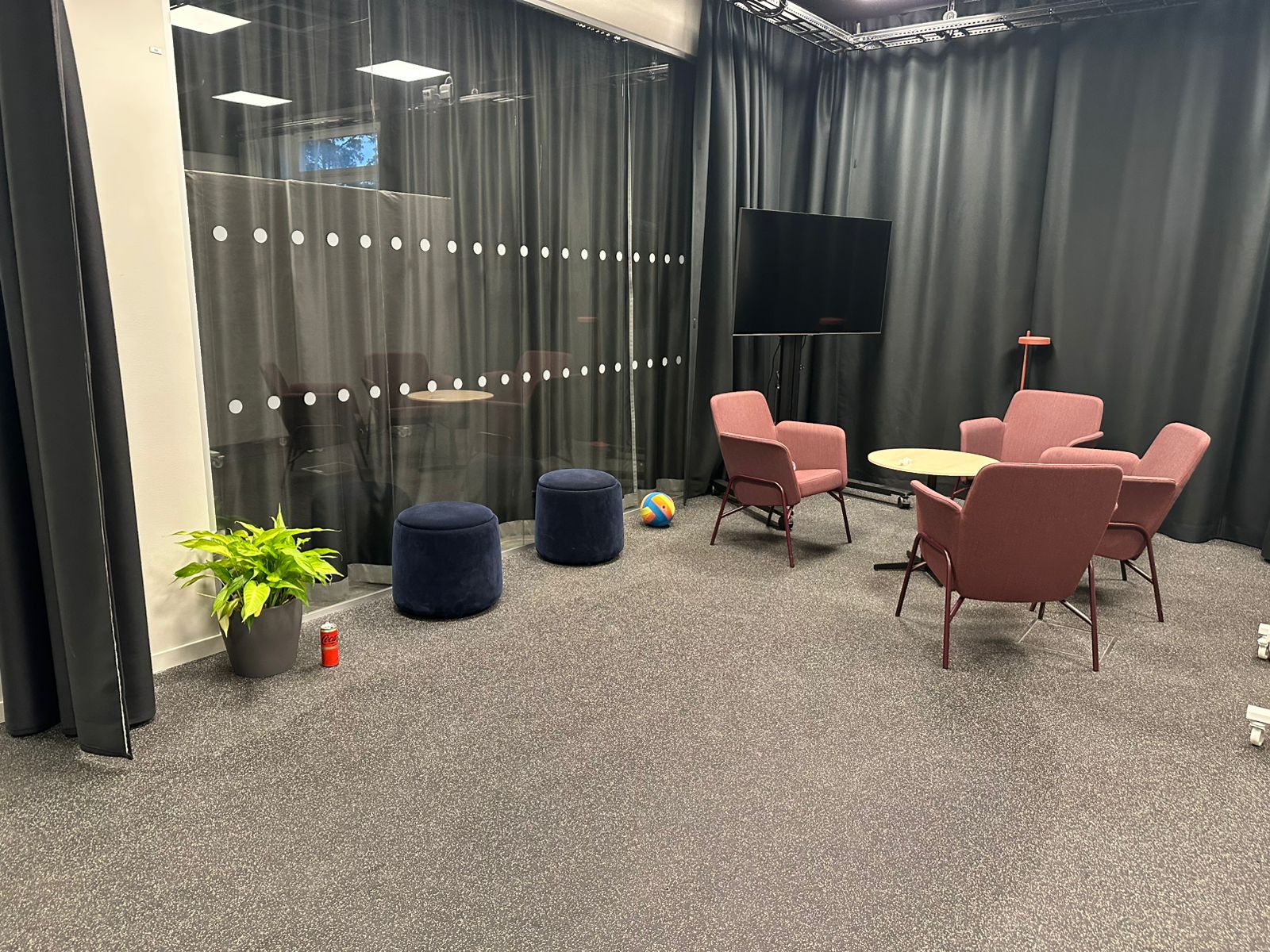}
        \vspace{-0.45cm}
        \caption{The office waiting room.}
        \vspace{0.15cm}
    \end{subfigure}
    \begin{subfigure}[t]{0.2\textwidth}
        \centering
        \includegraphics[width=\textwidth]{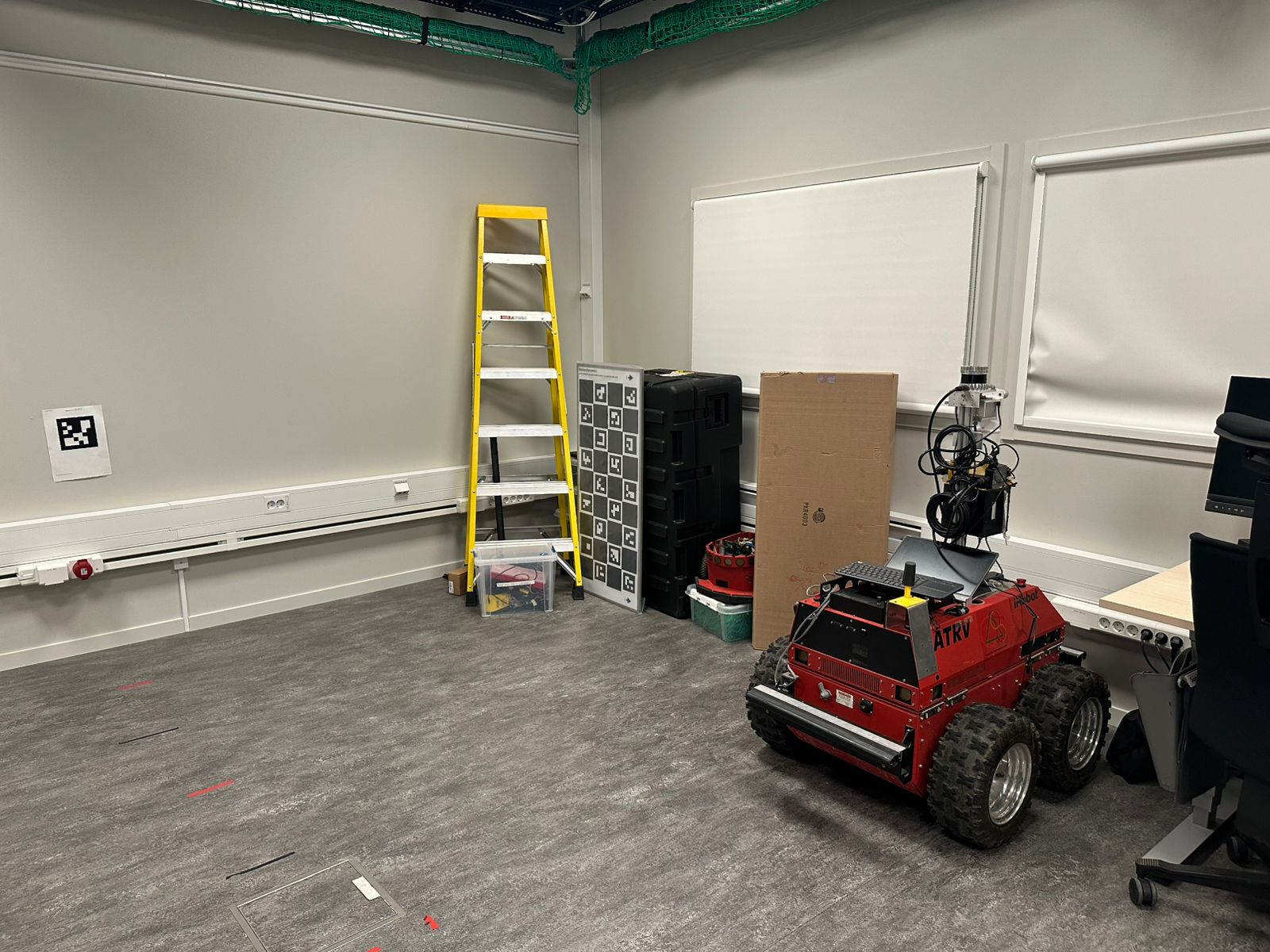}
        \vspace{-0.45cm}
        \caption{The robotics lab.
        }
        \vspace{0.15cm}
    \end{subfigure}
    \begin{subfigure}[t]{0.2\textwidth}
        \centering
        \includegraphics[width=\textwidth]{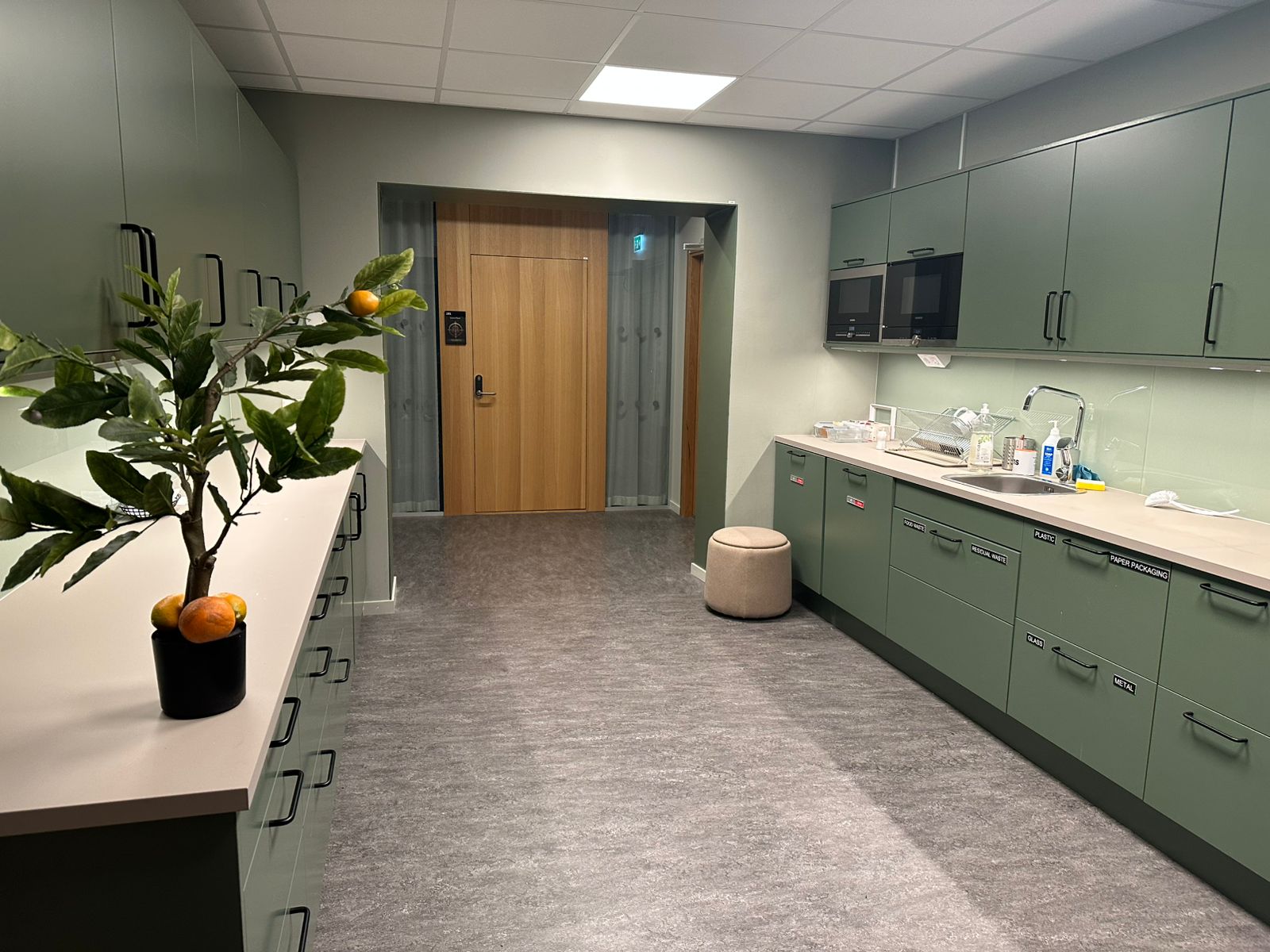}
        \vspace{-0.45cm}
        \caption{The kitchen area.}
    \end{subfigure}
    \begin{subfigure}[t]{0.2\textwidth}
        \centering
        \includegraphics[width=\textwidth]{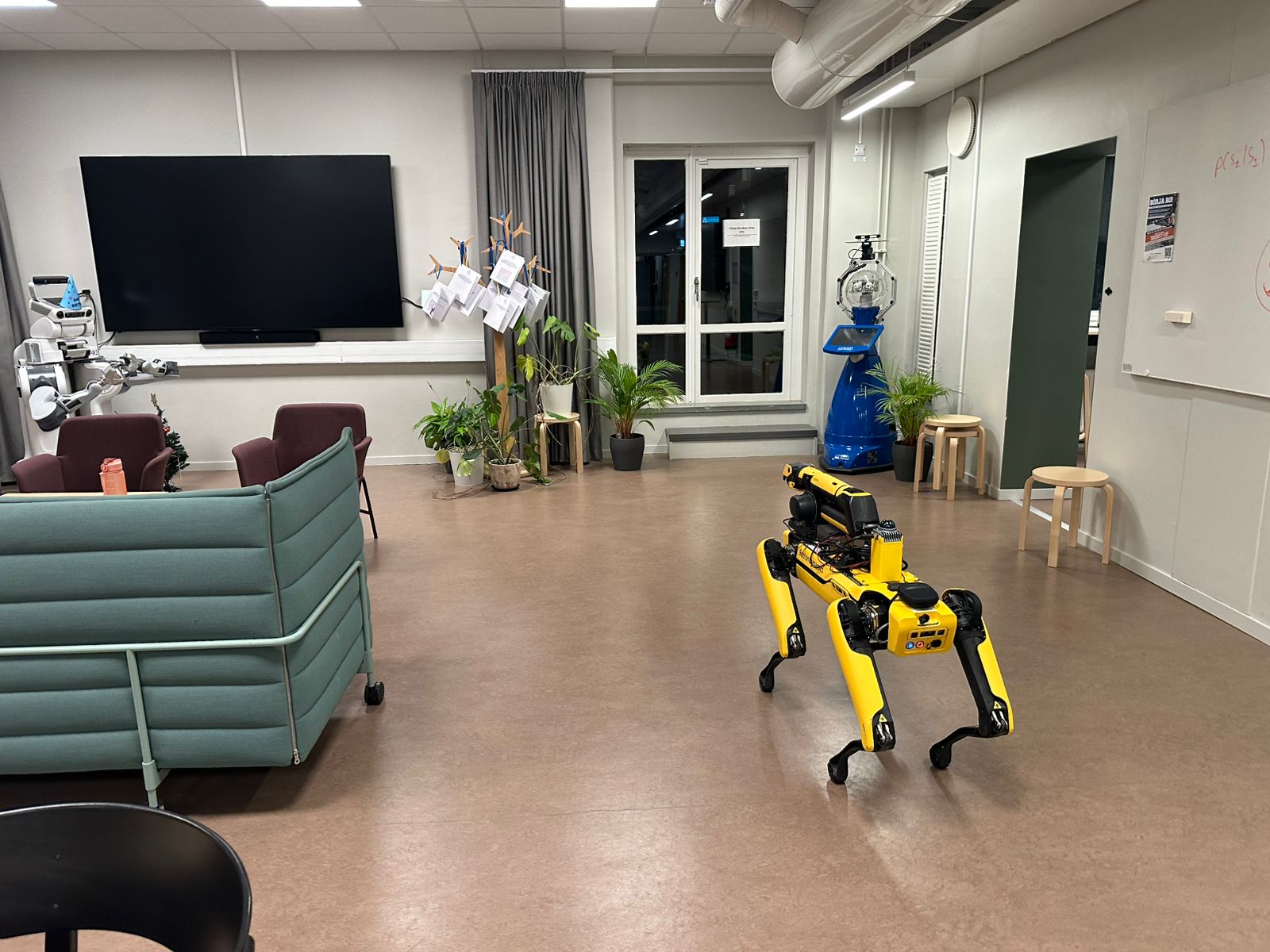}
        \vspace{-0.45cm}
        \caption{The office lounge.}
    \end{subfigure}
    \caption{Four real-world experiment scenes. We conducted 15 multi-object navigation episodes across these environments.}
    \label{fig:realworld}
\end{figure}

We conducted 15 multi-object navigation episodes across four realistic environments as described in the evaluation setup. The environments are shown in Fig.~\ref{fig:realworld}. Table~\ref{tab:real_world} compares our method against \cite{busch2025mapallrealtimeopenvocabulary} on real-world navigation tasks.

\begin{table}[h]
\centering
\vspace{6mm}
\begin{tabular}{l c | c c}
\toprule
Approach & SR\,/\,\# Found ($\uparrow$)  & FP ($\downarrow$) & TNF ($\downarrow$) \\
\midrule
OneMap\cite{busch2025mapallrealtimeopenvocabulary} & 53\%\,/\,24 & 40\% & 6.66\%\\
DIV-Nav (Ours) & \textbf{88\%}\,/\,\textbf{40} & 0\% & 11.11\% \\
\bottomrule
\end{tabular}
\caption{Comparison of real-world performance. Metrics: Success rate (SR) per object, number of objects found (\# Found), false positive rate (FP), and rate of attempts where the robot terminates the search without finding the object (TNF).}
\label{tab:real_world}
\vspace{-0.7cm}
\end{table}

Our system achieves an 88\% success rate, successfully finding 40 out of 45 total targets across all experimental runs. This represents a 66\% relative improvement over the OneMap baseline \cite{busch2025mapallrealtimeopenvocabulary} (and is statistically significant, 95\% Clopper-Pearson bounds).
We further report the false positive rate (FP), which shows that \cite{busch2025mapallrealtimeopenvocabulary} fails a substantial number of attempts due to misidentifying wrong objects, which we attribute to it not accounting for the spatial relationships. As a consequence, \cite{busch2025mapallrealtimeopenvocabulary} might call the FOUND action on objects that match the primary object, but do not satisfy the spatial constraints, leading to false positives.
We also observe significantly increased performance on real-world experiments for both methods, which we attribute to the challenges encountered in the benchmark that are not present in real-world.
Moreover, the performance gap between \cite{busch2025mapallrealtimeopenvocabulary} and DIV-Nav is larger in real-world experiments, since the proportion of queries containing spatial relationships is larger, which requires the system to possess capabilities to take spatial constraints into account.
The 11.11\% TNF rate has two contributing causes. First, LVLM validation occasionally rejects correct detections, as in the case of the red car in the robotics lab (Fig.~3b). Second, when a stated spatial relationship is factually unsatisfied in the environment, for instance when no television is present near any blue rug, $S_\mathrm{comb}$ (Eq.~1) never produces a high-scoring region satisfying both targets simultaneously. Note that while the maximum term in Eq.~1 still guides the robot toward the individually highest-scoring object, the subsequent LVLM validation will reject it as the spatial constraint remains unmet. Both cases result in a TNF outcome, and handling unsatisfiable spatial constraints remains an important direction for future work.

While the quantitative gains in Tables I and II represent the combined framework, our real-world experiments highlight the distinct role of the Intersection module in driving search efficiency. Specifically, by identifying regions of proximity where sub-targets likely co-exist, the system creates a focused search area that allows the robot to bypass isolated object instances. This is evidenced by the \textit{`birthday robot'} task, where the robot was directly guided towards the intersection region rather than naively validating every detected instance of a robot or tree in the environment. 

%

For one of the experiments in the environment depicted in Fig.~\ref{fig:realworld} (d), we provide the system with the query: \textit{`Find me the birthday robot with his Christmas tree'}.
In the experiment, the system correctly: (1) \textbf{Decomposes the query}: Identifies \textit{`robot'}, \textit{`Christmas tree'}, and their proximity relationship; (2) \textbf{Builds combined similarity map}: Creates intersection regions where both objects are likely to co-exist; and (3) \textbf{Validates spatially}: Confirms both the presence of individual objects and their spatial relationship through LVLM reasoning.

\begin{figure}[t]
    \centering
    \vspace{6mm}
    \includegraphics[width=0.4\textwidth]{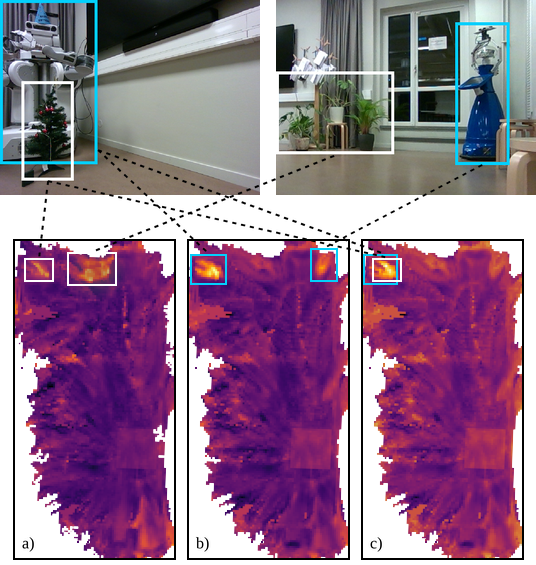}
    \vspace{-0.2cm}
    \caption{Data of a real world experiment: (top) RGB frames, (a) similarity map for query \textit{`plant'} , (b) similarity map for query \textit{`robot'} (c) and the resulting map of value intersection scores.}
    \vspace{-0.7cm}
    \label{fig:fig_4}
\end{figure}

To illustrate the capability of our value intersection scoring module to highlight specific object instances based on spatial proximity, Fig.~\ref{fig:fig_4} shows data collected during this experiment. We compute similarity maps for the queries: \textit{`plant'} and \textit{`robot'},  (Fig.~\ref{fig:fig_4} (a) and (b)) and the corresponding value intersection score (Fig.~\ref{fig:fig_4} (c)).

We identified two primary failure modes in our real-world experiments. First, \textbf{small or hidden objects}: Objects like a volleyball hidden in the waiting room or an electric kettle on a shelf failed to be detected due to limited visibility or small size in the semantic map's feature space. Second, \textbf{LVLM validation failures}: In some cases, objects with high similarity scores in the semantic map failed LVLM validation. For example, the red car in the robotics lab (Fig.~\ref{fig:realworld}b) was correctly mapped but incorrectly rejected during the visual validation step.


%% file: 06_conclusion.tex
\section{Conclusion}
In this work, we presented DIV-Nav, a multi-object navigation method that extends zero-shot navigation frameworks to handle spatially-constrained targets specified through natural language commands. By decomposing spatial queries into simpler semantic components via LVLMs and combining the resulting similarity maps, our approach enables robots to navigate to objects defined not just by what they are, but by where they are found relative to other objects.

Several limitations point to future directions. The semantic mapping struggles with small or occluded objects due to CLIP feature space limitations and map resolution constraints, VLM validation occasionally fails even when the semantic map correctly identifies target locations, and our approach is currently limited to proximity-based spatial relationships — generalizing to \textit{`far from'}, \textit{`between'}, or directional relationships would broaden applicability. Finally, integrating more advanced local planners to handle dynamic obstacles remains an avenue beyond the semantic reasoning framework proposed here.

%
%